# Predicting the Impact of Scope Changes on Project Cost and Schedule Using Machine Learning Techniques

Soheila Sadeghi [1]

*Abstract*— In the dynamic landscape of project management, scope changes are an inevitable reality that can significantly impact project performance. These changes, whether initiated by stakeholders, external factors, or internal project dynamics, can lead to cost overruns and schedule delays. Accurately predicting the consequences of these changes is crucial for effective project control and informed decision-making. This study aims to develop predictive models to estimate the impact of scope changes on project cost and schedule using machine learning techniques. The research utilizes a comprehensive dataset containing detailed information on project tasks, including the Work Breakdown Structure (WBS), task type, productivity rate, estimated cost, actual cost, duration, task dependencies, scope change magnitude, and scope change timing. Multiple machine learning models are developed and evaluated to predict the impact of scope changes on project cost and schedule. These models include Linear Regression, Decision Tree, Ridge Regression, Random Forest, Gradient Boosting, and XGBoost. The dataset is split into training and testing sets, and the models are trained using the preprocessed data. Model robustness and generalization are assessed using cross-validation techniques. To evaluate the performance of models, we use Mean Squared Error (MSE) and R2. Residual plots are generated to assess the goodness of fit and identify any patterns or outliers. Hyperparameter tuning is performed to optimize the XGBoost model and improve its predictive accuracy. The study identifies the most influential project attributes in determining the magnitude of cost and schedule deviations caused by scope modifications. It is identified that productivity rate, scope change magnitude, task dependencies, estimated cost, actual cost, duration, and specific WBS elements are powerful predictors.

## I. INTRODUCTION

IN construction projects, scope changes are a common problem that frequently lead to cost overruns and delays in the project schedule [1]. These changes may be requested due to internal factors, including differences in stakeholder perspectives, or external factors, such as unpredictable economic cycles, price fluctuations, highly competitive industries, and corruption [2]. In a project, scope is considered one aspect that can directly affects budget and timing. Despite its importance to project success, scope change appears to be one of the most neglected aspects of both agile and conventional project management [3], [4]. Due to escalating project demands and heightened expectations from stakeholders, project management professionals must enhance their proficiency in handling scope changes effectively. The growing intricacy of projects, along with stringent budget and timeline constraints, calls for a proactive and data-centric strategy for managing scope modifications. Furthermore, machine learning applications perform better than existing techniques, methods, and relying on human judgment on construction sites [5], [6].

Managing budgets effectively in project management is critical, particularly in environments characterized by uncertainty and resource constraints. Recent advancements, such as the S3RBAP framework highlighted in recent studies [7], underscore the importance of integrating risk management, resiliency, and sustainability principles into budget allocation processes, providing robust solutions to address uncertainties in project environments. Similarly, machine learning techniques have shown promise in project management applications, including cost estimation, risk assessment, and performance forecasting. Recent research on advanced time series methods like LSTM and ARIMA highlights ML's ability to capture dynamic project behaviors and integrate external factors to enhance predictive accuracy [8]. Building on these advancements, this study investigates the impact of scope changes on project cost and schedule using data-driven approaches to address complexities in dynamic construction environments.

The construction industry stands as a major contributor to the global economy, accounting for annual expenditures equivalent to a considerable 13% of the world's GDP [9]. Nevertheless, despite witnessing significant technological advancements in recent times, the construction sector continues to be regarded as the least efficient worldwide [10]. Harvard Business Review (2023) attributes this inefficiency to outdated technologies used in managing projects [11]. Gartner's research predicts that by 2030, 80% of project management tasks will be driven by artificial intelligence (AI), leveraging big data, machine learning (ML), and natural language processing (NLP) [12]. Consequently, project management professionals must adapt to these technological advancements by integrating AI and data-driven methodologies into their practices.

The construction industry's challenges are often linked to inadequate technological expertise and a low level of technology adoption. These issues have been associated with cost inefficiencies, project delays, subpar quality performance, uninformed decision-making, low productivity, and shortcomings in health and safety outcomes [12]. The present incapacity of the construction industry to fulfill the predicted infrastructure requirements in the near future can be attributed to its deficient adoption of digitalization and excessive dependence on manual approaches [13], [14].

Soheila Sadeghi is with the University of the Incarnate Word, San Antonio, TX 78209, USA (corresponding author, e-mail: ssadeghi@student.uiwtx.edu).

With the increasing availability of project data and advancements in data analytics, there is a growing opportunity to harness the power of data-driven techniques to predict the impact of scope changes on project performance more accurately. The ability to accurately predict the impact of scope changes is crucial for effective project management. It enables project managers to proactively assess risks, make informed decisions, and develop robust mitigation strategies. By leveraging historical project data and applying advanced analytics techniques, project managers can gain valuable insights into the likely consequences of scope changes [15]. This predictive capability empowers them to optimize resource allocation, prioritize tasks, and communicate potential risks to stakeholders more effectively. Moreover, integrating machine learning techniques in decision-making processes can mitigate potential negative impacts and optimize outcomes in complex environments [16].

Furthermore, a data-driven approach to scope change impact assessment can enhance project planning, as it provides a more objective and comprehensive understanding of the project ecosystem [17].

This study aims to develop a predictive model that utilizes project data to estimate the impact of scope changes on project cost and schedule. The predictive model will consider various project attributes such as the Work Breakdown Structure (WBS), task type, productivity rate, estimated cost, actual cost, duration, task dependencies, scope change magnitude, and scope change timing to capture the complex relationships and interactions that influence project performance.

The significance of this study lies in its potential to enhance project management practices by providing a data-driven approach to scope change impact assessment. Moreover, this study addresses the limitations of traditional scope change impact assessment methods, which often rely on subjective judgments and may not capture the full complexity of project dynamics. By incorporating a wide range of project attributes and utilizing machine learning algorithms, this approach aims to provide a more comprehensive and objective assessment of scope change impact, taking into account the intricate relationships and interdependencies within the project ecosystem.

The paper is structured as follows: Section II describes the dataset and the preprocessing steps undertaken to prepare the data for analysis. It includes dataset description, data preprocessing, and exploratory data analysis. Section III details the development of the predictive models, including Linear Regression, Decision Tree, Ridge Regression, Random Forest, Gradient Boosting, and XGBoost. It also covers the preprocessing and feature engineering steps, cross-validation process, residual plot analysis, and hyperparameter tuning. Section IV presents the results and discusses the models' performance, feature importance analysis, interpretation of model results, and practical implications for project management. Finally, Section V concludes the paper, highlighting the key findings, contributions to improved project planning and risk management, limitations, and future research directions.

## II. DATASET AND PREPROCESSING

### A. Dataset Description

The study utilizes a subset of the available dataset from historical records of a construction project provided by a civil contractor [18]. It contains detailed information on project tasks, including the Work Breakdown Structure (WBS), task type, estimated cost, actual cost, duration, task dependencies. The dataset also includes simulated values for key variables such as productivity rate, impact on cost, scope change magnitude, scope change timing, which were generated to represent realistic project scenarios and capture the variability commonly observed in real-world projects. Table I presents the descriptive statistics for the project dataset containing 221 rows. The table provides an overview of the key variables involved in predicting the impact of scope changes on project cost and schedule.

### B. Data Preprocessing

The dataset underwent several preprocessing steps to ensure data quality and prepare it for modeling. The column names were cleaned to maintain consistency, and missing values were checked. Categorical variables such as WBS, task type, and scope change timing were converted to factors to facilitate analysis.

Exploratory data analysis was performed to gain insights into the dataset. Summary statistics were calculated, and a correlation matrix was generated to identify relationships between numeric variables. The distribution of estimated cost was visualized using a histogram, excluding zero or near-zero values to avoid skewness. A scatter plot between estimated cost and actual cost was created to observe their relationship. Additionally, a box plot was used to examine the impact on cost across different scope change timing categories. Fig. 1 presents the histograms for various numerical columns in the dataset. The productivity rate displays a relatively normal distribution centered around a mean slightly above 1.0, indicating consistent productivity among resources. Both estimated and actual costs are right-skewed, suggesting that while most tasks have low costs, a few tasks incur significantly higher expenses. Task durations are predominantly short, with most tasks completed within 1-3 days, reflecting the common occurrence of brief tasks in projects. The impact on cost varies widely, with many scope changes causing minor cost impacts and a few resulting in substantial increases. Scope change magnitudes are generally small, indicating that large scope changes are infrequent but impactful when they do occur.

TABLE I
DESCRIPTIVE STATISTICS

| Statistic | Productivity_Rate | Estimated Cost | Actual_Cost | Duration | Impact_on_Cost | Scope_Change_Magnitude | Impact_on_Schedule |
|---|---|---|---|---|---|---|---|
| count | 221 | 221 | 221 | 221 | 221 | 221 | 221 |
| mean | 1.032 | 16121.55 | 16017.22 | 2.955 | 5374.94 | 0.173 | 0.570 |
| std | 0.081 | 8542.04 | 11013.61 | 1.174 | 5042.99 | 0.080 | 0.496 |
| min | 0.900 | 147.67 | 110.00 | 1.000 | 0.01 | 0.053 | 0.000 |
| 25% | 0.952 | 9337.87 | 8462.62 | 2.000 | 1773.42 | 0.100 | 0.000 |
| 50% | 1.027 | 15119.56 | 13301.40 | 3.000 | 4026.87 | 0.172 | 1.000 |
| 75% | 1.108 | 22639.34 | 21135.53 | 4.000 | 7491.26 | 0.240 | 1.000 |
| max | 1.197 | 30907.59 | 55980.80 | 5.000 | 24531.81 | 0.294 | 1.000 |

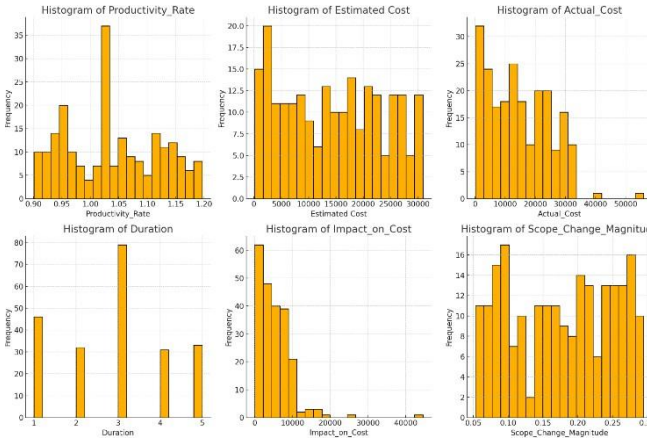

Fig. 1 Histograms for numerical variables

Fig. 2 presents scatter plots to examine the relationships between key variables. The plot of scope change magnitude versus impact on cost shows a weak positive correlation, suggesting that larger scope changes tend to increase costs, though other factors also influence this outcome. The relationship between scope change magnitude and impact on schedule is similarly weak, implying that while larger scope changes may affect the schedule, the correlation is not strong enough for precise predictions. Additionally, the plot of productivity rate versus impact on cost reveals no clear pattern, indicating that the productivity rate does not significantly correlate with cost impacts, highlighting the need to consider other influencing variables.

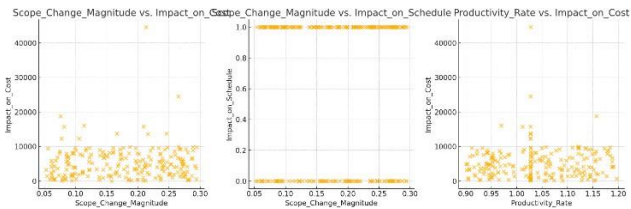

Fig. 2 Scatter Plots to Examine Relationships

To prepare the data for modeling, numeric features were scaled using the `preProcess` function from the `caret` package.

The dataset was then split into training and testing sets using an 80-20 ratio, with the `createDataPartition` function from the `caret` package ensuring reproducibility.

### III. PREDICTIVE MODEL DEVELOPMENT

#### A. Preprocessing and Feature Engineering

The preprocessing steps and model development approach aimed to create predictive models that can estimate the impact of scope changes on project cost and schedule. The data was prepared by filtering out columns with constant values and scaling the numeric features. Missing values were imputed to ensure a complete dataset for model training.*Machine Learning Models*

**Linear Regression:** A linear regression model was developed to predict the impact on cost using all available features in the preprocessed dataset. The 'lm' function from the 'stats' package was used to train the model on the training data. The model's performance was evaluated on the test set by calculating the mean squared error (MSE) between the predicted and actual impact on cost values.

**Decision Tree:** The 'rpart' function in the 'rpart' package was used to build a Decision Tree model to predict the impact on the schedule. A preprocessed dataset was used to train the model, with 'method' set to "class." The trained model was then used to make predictions on the test set, and its performance was evaluated using a confusion matrix generated by the 'confusionMatrix' function from the 'caret' package.

**Random Forest:** To estimate the impact on cost, the Random Forest function from the RandomForest package was used. A preprocessed dataset, with the tree number 100 as a parameter, was used to train the model. Following training, the model was used to predict test data and its performance was evaluated by calculating the mean squared error (MSE) and the R2 score.

**Ridge Regression:** Ridge Regression is a linear regression technique that introduces a regularization term to the ordinary least squares objective function. The regularization term, controlled by the parameter alpha, helps to prevent overfitting by shrinking the coefficients of less important features. In this study, the Ridge Regression model was trained using cross-validation with alpha set to 0. The model's performance was evaluated using Mean Squared Error (MSE) and the number of non-zero coefficients. This linear regression variant incorporates a regularization term into the ordinary least squares objective function. With Ridge Regression, the magnitude of the regression coefficients is constrained by regularizing the standard least squares objective function. During regularization, the hyperparameter alpha determines how strong the penalty will be applied to coefficients.

To evaluate Ridge Regression's performance without regularization, the alpha value was set to 0 in this study, so cross-validation can be used. Mean Squared Error (MSE) was used to measure the model's predictive accuracy, and the number of non-zero coefficients was examined to evaluate the model's ability to identify relevant predictors.

**Gradient Boosting:** This ensemble learning method combines multiple weak learners, typically decision trees, to create a strong predictive model. To improve prediction accuracy, the algorithm iteratively fits new models to minimize the residuals of previous models. In this study, Gradient Boosting models were trained using the 'gbm' function from the 'gbm' package. In this model, 100 trees are involved, the interaction depth is 3, and the distribution is Gaussian. After training, the model's performance was evaluated using MSE and R2 metrics on test data.

**XGBoost:** XGBoost (Extreme Gradient Boosting) is an optimized implementation of the Gradient Boosting algorithm. It incorporates several enhancements, such as regularization, parallel processing, and tree pruning, to improve model performance and scalability. In this study, an XGBoost model was trained using the 'xgb.train' function from the 'xgboost' package. The training data was converted to an 'xgb.DMatrix' object, and the model parameters were set with an objective of "reg:squarederror", a learning rate ('eta') of 0.1, and a maximum depth of 6. The model was trained for 100 rounds, and predictions were made on the test data. The model's performance was assessed using MSE and $R^2$ metrics.

*B. Cross-Validation*

To assess the robustness and generalization ability of the models, cross-validation was performed. The analysis involved preparing the data by filtering out columns with constant values and scaling the numeric features. After imputing missing values, multiple regression models were applied to predict the impact on the project schedule.

For the linear regression model, 10-fold cross-validation was conducted using the train function from the caret package. The results of the cross-validation, including the model's performance metrics, were printed for evaluation. The Ridge Regression model, with a Mean Squared Error (MSE) of 0.2362 and an R² value of 0.0281, explained only a small portion of the variance in the schedule impact.

Similarly, cross-validation was applied to the Decision Tree model using the train function with the method parameter set to "rpart" and the number of folds set to 10. The cross-validation results were printed to assess the model's performance across different subsets of the data. The Decision Tree model showed better performance with an MSE of 0.1984 and an R² value of 0.2098, indicating it could explain around 21% of the variance.

*C. Residual Plots*

Residual plots were generated for the XGBoost, Gradient Boosting, and Random Forest models to assess their performance and fit. The XGBoost residual plot showed tightly clustered residuals around the zero line, indicating minimal prediction errors and a good fit. There was no clear pattern in the residuals, suggesting that the model captured the underlying relationships well. Fig. 3 shows the residual plots for the XGBoost, Gradient Boosting, and Random Forest models. The residual plot for the XGBoost model depicts tightly clustered residuals around the zero line, indicating minimal prediction errors. The residual plot for the Gradient Boosting model shows a wider spread of residuals and a potential pattern, highlighting larger prediction errors and potential overfitting or underfitting issues. The residual plot for the Random Forest model demonstrates a relatively good fit with no clear pattern in the residuals, although some outliers are noted.

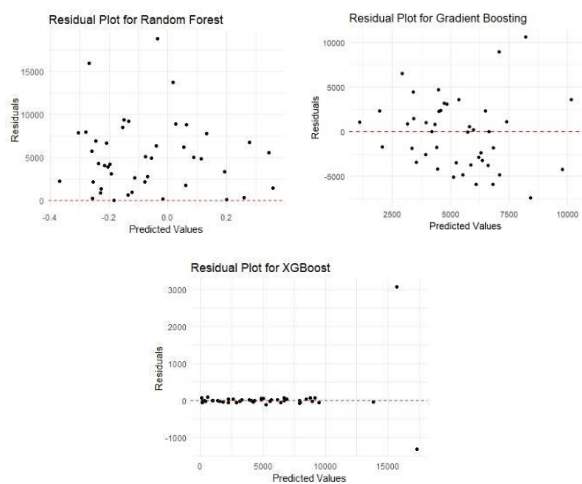

Fig. 3 Residual plots for the XGBoost, Gradient Boosting, and Random Forest models

The Random Forest model yielded an MSE of 0.2145 and an $R^2$ value of 0.1353, while the Gradient Boosting model had an MSE of 0.2447 and an $R^2$ value of 0.0455. The XGBoost model performed relatively well with an MSE of 0.2094 and an $R^2$ value of 0.1804.

The Gradient Boosting residual plot exhibited a wider spread of residuals and a potential pattern, indicating larger prediction errors and possible overfitting or underfitting issues. The Random Forest residual plot demonstrated a relatively good fit with no clear pattern in the residuals, although a few outliers were present.

*D. Hyperparameter*

The cross-validation results revealed the best set of hyperparameters for the XGBoost model based on the lowest RMSE. Fig. 4 shows the performance metrics (RMSE, $R^2$, MAE) across different cp values for the XGBoost model. The RMSE values across different cp values, indicating the optimal cp value for minimizing RMSE in the model. The $R^2$ values across different cp values, providing insights into the model's explanatory power for different complexity parameters. The Mean Absolute Error (MAE) values across different cp values, aiding in the selection of the cp value that minimizes prediction errors.

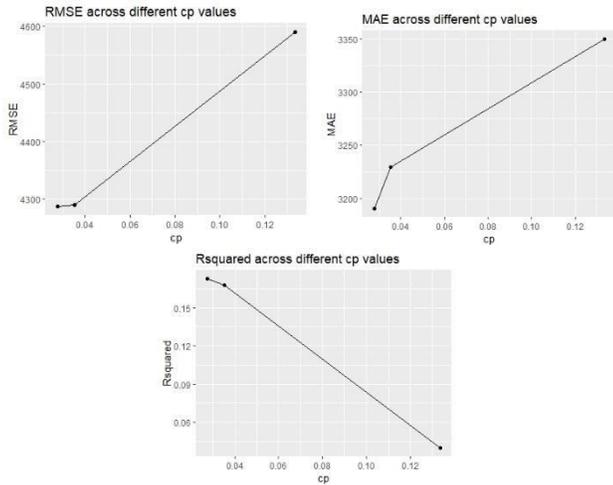

Fig. 4 Performance metrics (RMSE, R-squared, MAE) across different cp values for the XGBoost model

The selected hyperparameters included a learning rate (eta) of 0.1, a maximum tree depth (max_depth) of 3, a minimum loss reduction (gamma) of 0, a subsample ratio of columns (colsample_bytree) of 0.7, a minimum sum of instance weight (min_child_weight) of 1, a subsample ratio of training instances (subsample) of 0.7, and a number of boosting rounds (nrounds) of 100. These hyperparameters achieved an RMSE of 0.2727304, indicating good performance on the validation sets.

*E. Prediction Interpretation*

The predictions obtained from the trained model represent the predicted impacts on cost for each corresponding row in the new_data dataset. The prediction values ranged from negative to positive, indicating potential increases and decreases in cost impact. The magnitude of each prediction signified the relative impact on cost, with larger absolute values suggesting a greater impact.

*F. Feature Importance*

In addition to the main analysis, a separate feature importance assessment was conducted using the XGBoost model trained without actual costs. This analysis is particularly relevant in scenarios where the project is ongoing or in its initial stages, and actual cost data is not yet available. By excluding actual costs from the model, we can gain insights into the relative importance of other project features in predicting the impact on cost.

The exclusion of actual costs from the model simulates a situation where project managers need to make cost impact predictions based on available information such as the Work Breakdown Structure (WBS), scope change magnitude, task dependencies, and other relevant factors. This analysis provides valuable insights for proactive cost management and decision-making in the absence of actual cost data.

The feature importance plot generated for the XGBoost model trained without actual costs revealed the relative significance of different project attributes. The most influential feature was wbs.86, likely representing a critical element within the WBS. This suggests that specific components or tasks within the project structure have a substantial impact on overall costs, even when actual cost information is not considered.

Other notable features included scope_change_magnitude, highlighting the importance of considering the extent of scope changes in predicting cost impact. Task dependencies (e.g., task_dependencies51) and specific WBS elements (e.g., wbs.SWPC, wbs.CSA, wbs.62) also emerged as relevant factors, indicating their influence on project costs.

These findings align with fundamental project management principles, emphasizing the significance of carefully planning and monitoring critical project components, managing scope changes effectively, and considering task dependencies to control costs. By focusing on these key aspects, project managers can proactively identify potential cost drivers and take preventive measures, even in the absence of actual cost data.

## IV. RESULTS AND DISCUSSION

*A. Feature Importance Analysis*

The feature importance analysis from the XGBoost model highlighted that productivity rate, scope change magnitude, task dependencies, estimated cost, actual cost, duration, and WBS were the most significant factors affecting the project cost and schedule. This insight can guide project managers in focusing on these critical areas to mitigate cost overruns and schedule delays.

The productivity rate emerged as the most influential feature, emphasizing the importance of monitoring and optimizing productivity to control project costs and timelines. Scope change magnitude and task dependencies also played crucial roles, indicating the need for effective scope management and careful consideration of interdependencies among project tasks.

Estimated cost, actual cost, and duration were identified as significant predictors, highlighting the importance of accurate cost estimation, cost tracking, and schedule management. The analysis also revealed the relevance of specific WBS elements, suggesting that certain project components have a greater impact on cost and schedule performance. Fig. 5 presents a correlation matrix heatmap that underscores the relationships between different project metrics. A high correlation between estimated and actual costs (0.94) is expected, indicating these variables move closely together. The mutually exclusive categories of `Type_L (labor)` and `Type_M (machinery)` have a strong negative correlation (-1.0). Scope change magnitude shows moderate correlations with both impact on cost (0.57) and impact on schedule (0.37), suggesting it somewhat influences these impacts. However, most other variables exhibit low correlations with the performance metrics, indicating no single feature strongly predicts these impacts.

This emphasizes the complexity of predicting project performance, necessitating the consideration of multiple factors and potentially more advanced modeling techniques.

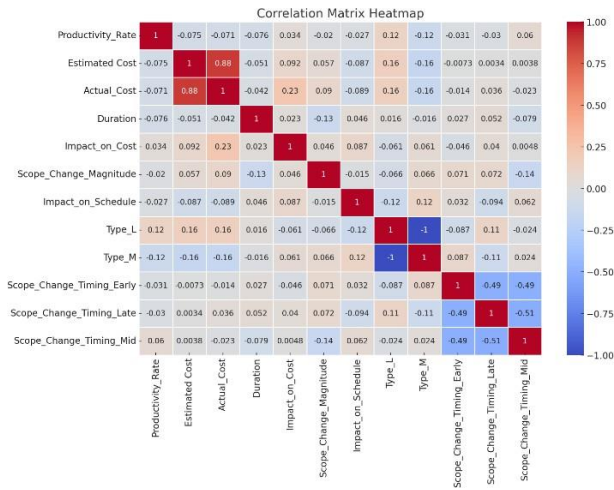

Fig. 5 Correlation matrix heatmap

### B. Interpretation of Model Results

The XGBoost model demonstrated excellent performance in predicting the impact on cost, with an $R^2$ value of 0.985. This indicates that the XGBoost model captured the underlying patterns in the data very well and provided highly accurate predictions for the impact on cost.

The Ridge Regression model also achieved good results, with an $R^2$ value of 0.991. The model retained 66 non-zero coefficients, suggesting that it effectively captured the important features while regularizing less significant ones.

The Decision Tree model outperformed other models in predicting the schedule impact, as indicated by its lowest MSE and highest $R^2$ value. The feature importance analysis from the XGBoost model highlighted that productivity rate, scope change magnitude, task dependencies, estimated cost, actual cost, duration, and WBS were the most significant factors affecting the project schedule. This insight can guide project managers in focusing on these critical areas to mitigate schedule delays. Fig. 6 shows the feature importance for schedule impact, emphasizing that productivity rate, scope change magnitude, and task dependencies are the most critical factors influencing the project schedule.

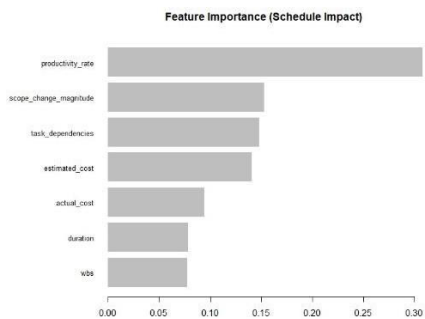

Fig. 6 Feature importance for schedule impact

In contrast, the Gradient Boosting model yielded an $R^2$ value of 0.115, indicating room for further optimization to enhance its predictive power. The Random Forest model also showed relatively higher MSE values and lower $R^2$ scores compared to XGBoost and Ridge Regression.

### C. Implications for Project Management

The machine learning-based predictive models developed in this research offer project managers a powerful tool to forecast the likely impact of scope changes on project budget and timeline. By leveraging vast unused project data to train ML models, project managers can gain insights and generate realistic forecasts. Capturing comprehensive, diverse data across projects is crucial for accurate ML predictions. Adopting ML can improve construction planning, scheduling, and forecasting accuracy and efficiency. Project managers can use these ML predictions in planning and resource allocation to reduce delays and shortages. The feature importance analysis serves as a valuable compass, guiding project managers to focus their attention on the most influential drivers of project success, such as productivity metrics, scope change management, task dependencies, and critical WBS components. By proactively monitoring and addressing these factors, project managers can identify and tackle potential roadblocks before they derail the project's financial and temporal objectives.

Moreover, the models' ability to predict the impact of scope changes enables project managers to conduct scenario analysis and assess the potential consequences of proposed changes. This information can facilitate effective communication with stakeholders, justify decisions, and secure necessary resources to mitigate the impact of scope changes.

## V. CONCLUSION

This study developed a data-driven and ML-based approach to predict the impact of scope changes on project cost and schedule. Multiple machine learning models, including Linear Regression, Decision Tree, Ridge Regression, Random Forest, Gradient Boosting, and XGBoost, were developed and evaluated to predict the impact of scope changes on project cost and schedule. The findings confirmed that the XGBoost model demonstrates the best performance in predicting the impact on cost, while the Decision Tree model outperforms others in predicting the impact on schedule.

Moreover, feature importance analysis provided valuable explanatory insights into the contributions of the input features to the cost and schedule impact prediction. It was identified that the most influential parameters on project cost and schedule are productivity rate, scope change magnitude, task dependencies, estimated cost, actual cost, duration, and specific WBS elements.

This study presents an effective example of using machine learning techniques to improve cost and schedule impact prediction accuracy in projects. In theory, the findings of this study provide new insights into the impact of scope changes on project cost and duration. One of the key barriers to effectively managing scope changes in projects is the lack of accurate predictions of their impact on cost and schedule. Thus, the understanding gained through the findings of this study may help project managers to proactively assess and mitigate the consequences of scope changes, leading to improved project performance and success.

Despite the contributions, there are some limitations, and hence the following recommendations are made for future studies. First, the models could be further validated with additional data sets when they become available. Future research can explore the integration of additional project data, such as resource allocation, risk factors, and external influences, to further enhance the predictive capabilities of the models. Investigating the applicability of the developed models across different project domains and industries can also provide valuable insights into the generalizability of the findings.

Then, the models could be advanced to predict the probability and magnitude of cost and schedule overruns. At any stage of the project, the advanced models could predict the probability and magnitude of overruns at completion, helping to spot risky projects early. Finally, future studies may also investigate feature interaction in addition to feature importance within the models. Understanding how different project attributes interact and influence each other can provide deeper insights into the complex dynamics of scope changes and their impact on project performance.